# Low Cost Semi-Autonomous Agricultural Robots In Pakistan-Vision Based Navigation

Scalable methodology for wheat harvesting

Muhammad Zubair Ahmad, Ayyaz Akhtar, Abdul Qadeer Khan, Dr.Amir Ali Khan, Dr.Muhammad Murtaza Khan
Department of Electrical Engineering, School of Electrical Engineering and Computer Science
National University of Sciences and Technology, NUST
Islamabad, Pakistan
09beezahmad@seecs.edu.pk, 09beeakhtar@seecs.edu.pk, 09beeaqkhan@seecs.edu.pk, amir.ali@seecs.edu.pk,
Muhammad.murtaza@seecs.edu.pk

*Robots have revolutionized our way of life in recent years. One of the domains that has not yet completely benefited from the robotic automation is the agricultural sector. Agricultural Robotics should complement humans in the arduous tasks during different subdomains of this sector. Extensive research in "Agribotics" has been carried out in Japan, USA, Australia and Germany focusing mainly on the heavy agricultural machinery. Pakistan is an agricultural rich country and its economy and food security are closely tied with agriculture in general and wheat in particular. However, agricultural research in Pakistan is still carried out using the conventional methodologies. This paper is an attempt to trigger the research in this modern domain so that we can benefit from cost effective and resource efficient autonomous agricultural methodologies. This paper focuses on a scalable low cost semi-autonomous technique for wheat harvest which primarily focuses on the farmers with small land holdings. The main focus will be on the vision part of the navigation system deployed by the proposed robot.*

## I. Introduction

Agriculture is the backbone of Pakistan's economy. It contributes 21% to the GDP and employs 45% of the total labor force. An estimated 60% of the rural population earns its livelihood from this sector [17]. Wheat is an important cash-crop and is also the staple food of the country. One of the major handicaps in the progress of agricultural sector is the scarcity of labor, caused by rural urban migration. According to a study of the Islamabad district, 6-14% of the total migrants are from landless agricultural labor and 7-33% from tenants [18]. The severe weather conditions during the harvesting season (mid-April to mid-May) and prolonged exposure to UV radiation causes fatigue, cataracts and can lead to skin cancer. Moreover, the harvested process needs to be expedited because of the risk of floods. These calamities are equally damaging to harvested crop because an average farmer in Pakistan lacks access to storage facilities. In Pakistan, 86% of the farmers have landholdings under 5 hectares with an average yield of 2.707 metric tons per hectares [17] which does not translate to a sizable profit due to expensive labor, fuel, fertilizers and pesticides, fertilizers and pesticides.

Recent era has seen huge developments in the information technology, machine vision and robotics communities. The need of the hour is to exploit these technologies for the local agricultural industry. With a national focus, the developed solutions should be cost effective, scalable and efficient. The goal should be to complement the farmers as and when desired. Indigenous development of such solutions should eventually result in immediate deploy ability, trouble shooting and up gradation.

Agricultural automation is a concept which dates back to the 1920's [1]. However, the initial progress was relatively slow and first fully driverless systems using leader cables were developed in 1979 [2]. While, the development of the platform is fundamental to these autonomous farm vehicles, it is the sensor part that is critical to the correct navigation of these machines. In this regard, machine vision algorithms for tractor guidance were developed from theoretical perspectives in 1984 [3] and 1985 [4] with their implementations on tractors taking place in 1995 [5] and 1997 [6].

Near Infrared (NIR) images and Bayes classifiers were used to separate out soil and crop in a vision based guidance system in 1987 [7]. In 1996, a method based on the separation of cut/uncut crop was developed [8] and it successfully logged 40 hectares of land [9]. Carrier-phase GPS and real-time kinematic (RTK) GPS [11] have also been used for autonomous navigation. Other sophisticated systems include development of vision based algorithms for vehicle navigation based on crop conditions, e.g., tilled/untilled soil boundary [12] and navigation via sensor fusion.

The recent works include the tomato picking robot which localizes on the basis edge information and morphological operations. It further incorporates stereovision for depth information. It yielded an accuracy of 93.3% in non-overlapping conditions and 84% in overlapping conditions [4]. Another similar work is radicchio harvesting where the localization is done based on the thresh-holding in HSL domain and morphological operations to locate the center of the plant [1]. Variable field of view based guidance system was used for navigation in the corn field. The variable field of view, the near, far and lateral was generated by controlling yaw and pitch of the camera. The guidance lines made using the morphological operations were then navigated using fuzzy logic [5].

State of the art in the motion planning domain is the development of extended kinematic models which incorporate the slippage [2]. To make cost effective routing and motion planning so that spatiotemporal constraints of the co-operating vehicles, e.g. the relative position of combine and the transporter so that combine tank does not fill up and transporter travels minimum distance while doing this, are incorporated for better optimization [3].

While most of the afore-mentioned methods are effective in their domain of application, they are generally expensive. Moreover, most of them are developed for crops that are green and planted in straight lines and thus not well suited to our problem. In our local environment, the major challenge is presented by the random arrangement of wheat plants and its similarity in color to soil especially under bright sunlight. We propose a cheap solution using standard camera with an assisted GPS as sensors, followed by processing of the images for crop segmentation purposes. The resultant control signals are then passed to the actuator for controlling the platform. The proposed method applies image transformation on the acquired image and performs image segmentation by using cut planes and thresh holding and uses morphological operations and assisted-GPS for further localization. The method is discussed in detail in the following section.

II. PROPOSED METHODOLOGY

The proposed system is presented in Fig 1. The overall system can be sub-divided into four main modules. The first one is the sensor module that incorporates a standard camera along with an assisted-GPS unit. We utilized the camera and the assisted-GPS units of a normal mobile phone thus avoiding over the top costs. The image acquisition rate is a parameter that can be adjusted before harvesting and depends on the local terrain and the speed of the vehicle. The assisted-GPS offers a spatial resolution of **0.5-1**m. The purpose of obtaining the GPS co-ordinates is to obtain an idea of the position of the autonomous platform as well to mark the boundary location of the field to be harvested.

The second main module is the image processing module which takes its input from the sensor module. The resulting image is processed in order to segment out the uncut crops from the soil and the cut crops. The algorithmic details will be discussed later in this section. The results of this processing unit are passed to a controller (actuator) system, which takes the necessary decision for navigating the platform through motor controls. This unit is devised using a microcontroller and serves principally to control the steering and drive motors which form the fourth and final module of the system.

The main module of the system is the processing module and we will focus on some of the details of this module in this paper. The remaining modules involving the mechanical fabrication and electrical design of the proposed system will be discussed only briefly in a later section.

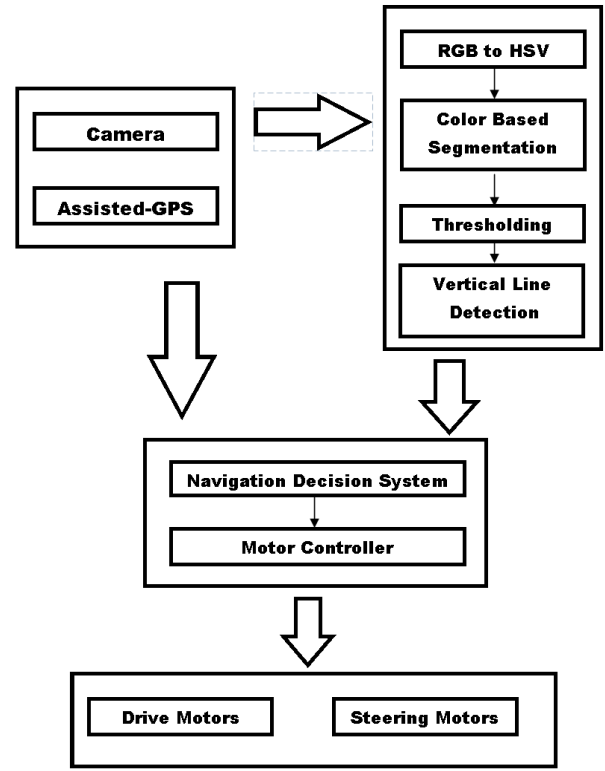

**Fig 1: Detailed System diagram showing different modules of the proposed system**

Once the wheat (our target crop) is ripe, it changes its color from golden to yellow. Soil having a similar color, during daytime, the presence of bright sunlight and the lack of moisture in the soil make the contrast in color indistinguishable. In order to address this scenario, we need to develop some method to enhance the contrast between the ripe crop and the soil. The goal will be to project the data from RGB space onto a space that offers more dissimilarity between the soil and the ripe crop.

HSV (hue-saturation-value) color space is more suited to our application than the RGB color space as it separates the effect of chrominance and luminance, thus theoretically reducing the effect of the bright sunlight. The HSV color space can be represented in cylindrical coordinates in contrast to the representation of RGB color space in rectangular coordinates. The Hue map corresponds to the variation in the colors, while the saturation corresponds to the intensity of the color, and the Value represents the amount of black color being mixed. Hence to segment out crop from the soil, at the first stage we only extract the portions in the yellow region which separates the soil and crop from the background. We segment out the yellow color from the HSV cylinder on the basis of following cut planes.

$$cutting\ plane\ 1: Hue = \varphi_1 \quad (1)$$

$$cutting\ plane\ 2: Hue = \varphi_2 \quad (2)$$

$$a * Saturation + Value = b \quad (3)$$

$$\theta = \varphi_2 - \varphi_1 \quad (4)$$

$\Phi1$ and $\phi2$ are the values of the $\Theta$, from the cylindrical coordinates. These are half planes originating from the center of the cylinder and $\Delta\Theta$ from Eq. 4 represents the angle between these planes while the parameters a, b vary the third plane in saturation and value plane thus changing the shades of the required color we segment out.

Eqs. 1-3 represent the three cut planes while the fourth equation '$\theta$' is a representation of the range of true colors which we segment out (in our case yellow). After applying this transformation, and considering Eq. 4, we only have soil and crop pixels that dominate. The crops are separated out from the soil by thresh-holding this true color range. The threshold was chosen empirically. This threshold parameter will have to be slightly tuned in at the time of deployment of the farm vehicle and will depend on the soil and crop conditions on ground. This parameter selection can be automated in some other work, e.g., using a database of the possible soil crop contrasts at different terrains and during different times.

While, a reasonable amount of crop segmentation is achieved by color based segmentation, the result does not extract the crops uniquely. Other factors such as the presence of residual crop (part of the harvested stalk still standing in the soil), harvested crop (which is dumped in the field to be collected later), and the similarity in crop-soil color due to lighting effects also manifest themselves in the result. This necessitates the use of some complementary features specific to the crops which can aid us in their extraction. One such feature is the texture of crop. The simplest texture of the crop is its vertical stalk which can distinguish it from the other surrounding factors. Vertical line detection is thus performed on the thresh-holded image to isolate the crops from the soil. In the vertical line detection, a variable tilt tolerance of '$x$' degrees is used to cater for windy conditions and natural tilt, where the parameter '$x$' will have to be tuned in during initial setup. Roughly, this parameter falls in the range of -5 to +5 degrees.

Although, at this stage, the crop has been separated out from the soil, as well as from most of the cut crop lying on the ground, we cannot distinguish with complete certitude between the residual cut crop and the uncut crop. To detect the crop that has been cut (and may have some residual part), we apply a simple check on the lengths of the detected segments. If the length of a detected segment is below a certain thresh-hold, we associate it to the residual cut crop (remaining part of the harvested stalk still standing in the soil), hence discard it. While most of the harvested crop dumped in the field, being horizontally oriented, has been separated out during vertical line detection, a check on the length of the line is also applied to prevent the effects of noisy environment.

There are certain parameters, like 'a' and 'b' in Eq. 3 that need to be set manually at the start of the session but also during the session depending on the variation of solar intensity. As an indicator, during a normal day (ruling out the extreme weather), these parameters have to be tuned about four times from 9:00 am to 4:00 pm. This information on the crops to be harvested is passed onto the navigation platform.

We next focus our attention to the navigation part of the robot. Fig. 2 describes the approach of the robot to the separation between two fields. Once, the first field has been cut, robot should be able to identify the end of field.

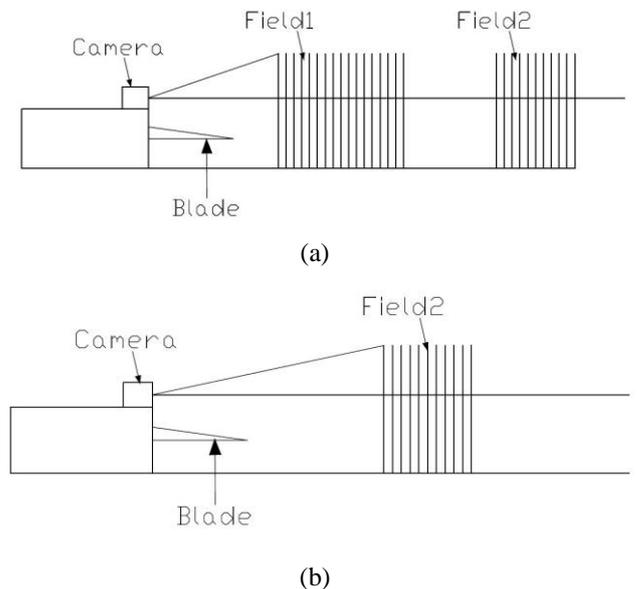

(a)

(b)

**Fig. 2:** This figure demonstrates how robot will approach the end of field (a): It shows the robot in field 1 progressing towards field 2. (b): This figure shows how the robot has now harvested field 1 and reached the separation between fields.

This end of field detection is based on the assumption that the wheat plants within a field are densely packed and the only breaks that occur are at the boundaries as in Fig. 2. The wheat plants are roughly the same height so they translate to the same height in the images acquired. From the behavior of a pinhole camera we know that the faraway objects appear smaller in the camera, so the plants that are distant will appear at a lower height in the image. This effect is demonstrated in Fig 3 using the basic optics. As we can only see the plants directly in front of us so in a continuous field all crops will be at the same distance from the camera. When these crops have been cut the crops in the adjacent field will be visible. A gap between two fields will translate into a sudden height difference in the acquired image.

When an end of plot is detected we take a right turn and continue till the end of plot. We have assumed that the field will be rectangular and robot will be placed at the lower right corner. In this way we reduce our control algorithm to just the straight motion and a right turn.

Assuming a rectangular geometrical boundary for the field, once an end of field is detected, we take a right turn and continue till the end of field in the other direction. Initially, the robot will be placed at the lower right corner. In this way we simplify our control algorithm to follow just the straight trajectory and then a right turn.

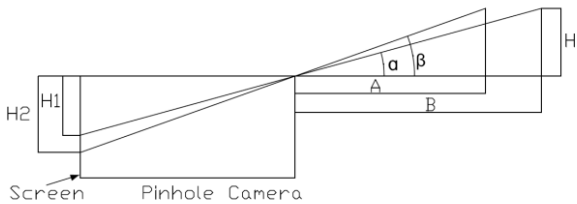

**Fig 3: The diagram shows how two objects of same height h subtend different angles at the pinhole with the closer object subtending the larger angle thus forming a larger image on the camera screen.**

The mechanical assembly of the robot consisted of a four wheel design which consisted of the two rear drive wheels. There were two rear drive motors and a steering motor mounted at the front. The signals to these were generated via an arduino microcontroller on the basis of the processing of the sensory data.

III. RESULTS AND DISCUSSION

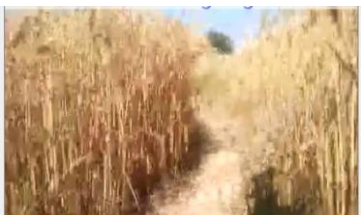

(a)

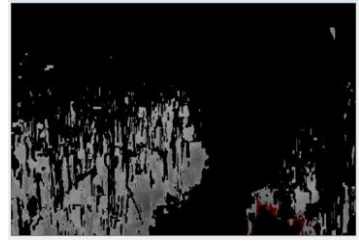

(b)

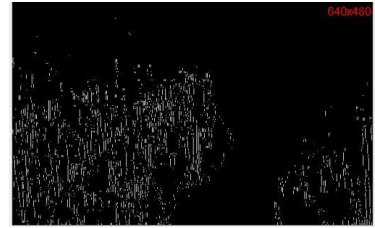

(c)

**Fig 4: (a) Original image captured in a field featuring a separation between two fields (b) The results after the color-segmentation and thresh-holding (c) The vertical lines detected in the image**

In this section, we demonstrate the results of the main image processing unit and its capability to segment out the crops. The results of the navigation part are not presented here as their illustration goes beyond the scope of this paper. Fig. 4(a) shows the image acquired from a ripe wheat field in bright sunlight. This case presents a scenario when we have the crop, soil intersection and would be used to demonstrate the separation between the two. The effects of shadows are also pretty much noticeable in this image. Fig. 4(b) shows the results of performing the HSV transformation and the associated thresh-holding on the cut plane. As can be seen, the background is very well separated out by this method. The application of vertical line detection gives us a clear mapping of the ripe crops as presented in Fig. 4 (c). The effect of the shadow is dominant near the base but it is nullified by the fact that wheat pixels will be connected to the ground pixels at the base of the plant.

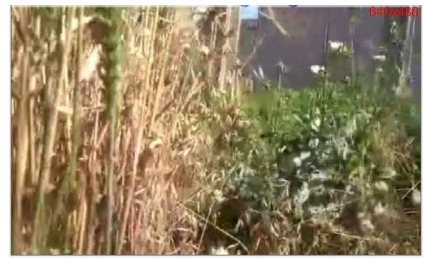

(a)

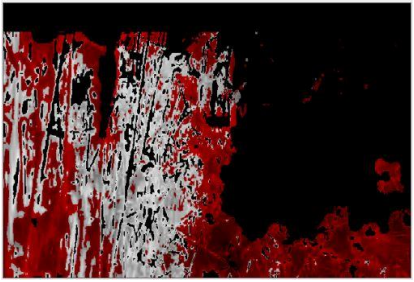

(b)

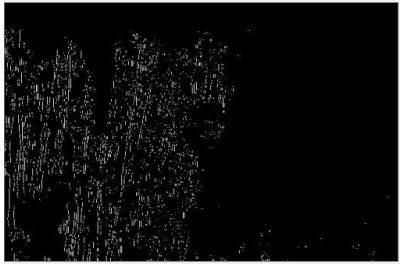

(c)

Fig 5: (a) Original image showing wheat plants along with weeds (b) The results after the color-segmentation and thresh-holding (c) The vertical lines detected in the image

Fig 5 presents another scenario when the robotic platform is closer to the corner of a field where there are a lot of weed and some unripe wheat plants. As it can be seen in Fig. 5(b-c), the proposed algorithm accurately segregates out only the ripe wheat plants from the image. We calculate the centroid of the segmented portion and aim to keep the centroid align with the center of the screen. In the straight motion the control signals are generated by this method.

The robotic platform was tested on other scenarios where it also offered very satisfactory results. Although, there are certain parameters that are required to be tuned at the start of the session, they are not handicapping. The automatic tuning of these parameters will be taken up in a later work.

Thus using a mobile phone quality camera mounted on the platform along with the assisted-GPS allows to control the navigation of the robot through a microcontroller. The goal of this paper was to give a proof of concept, and a standard laptop, placed on the robot was used for computational purposes. The concept can be extended to the industrial scale with embedded systems for processing along with an industrial scale mechanical design.

The control techniques that were not discussed in this paper will have to be adapted for different platforms. We designed and tested the controls for a small autonomous model of a power reaper. We used a model with a steering control and two rear wheel drive motors. It was just to demonstrate the concept and was not designed for real field applications. As discussed earlier, an industrial grade system can be implemented as an extension of this project.

## IV. CONCLUSIONS

In this paper, we propose an algorithm for automated vision based navigation of an agricultural robot. The proposed solution is scalable as it is platform (robot) independent. The vision algorithm is very simplistic and thus computationally efficient for real time deployment. It was demonstrated that suitable navigation can be achieved through the proposed system. Moreover, it is developed based only on a mobile camera. The estimated cost of the system that we added on to the test vehicle was about USD 200. The proposed system can be developed as an add-on component to be installed with multiple platforms. Parameter training can be automated by training on a dataset of images in different weather conditions. The need for a separate computer can be eliminated by utilizing image processing software for mobile operating systems. Low cost automated reaper models can be developed to cater for the needs of small scale farmers. Swarm robotics can also be used to make small robots cooperate to perform large scale tasks efficiently and effectively.

## V. REFERENCES

[1] M. M. Foglia, G. Reina, "Agricultural Robot for Radicchio Harvesting," *Journal of Field Robotics,* vol. 23, no. 6, p. 363–377, 2006.

[2] R. Lenain, B. Thuilot, C. Cariou, P. Martinet, "High accuracy path tracking for vehicles in presence of sliding: Application to farm vehicle automatic guidance for agricultural tasks," *Auton Robot,* vol. 21, pp. 79-97, 2006.

[3] S. Scheuren, S. Stiene, R. Hartanto, J. Hertzberg, M. Reinecke, "Spatio-Temporally Constrained Planning for Cooperative Vehicles in a Harvesting Scenario," *Künstliche Intelligenz,* vol. 27, 2013.

[4] J. Li, S. Chen, Y. Chen, Y. Chiu, W. Tu, P. Pan, "Study On Machine Vision System For Tomato Picking Robot," in *Proceedings of the 5th International Symposium on Machinery and Mechatronics for Agriculture and Biosystems Engineering (ISMAB*, Fukuoka, Japan, 2010.

[5] J. Xue, Z. Xhang, T. E. Griffit, "Variable field-of-view machine vision based row guidance of an agricultural robot," *Computer and Electronics in Agriculture,* vol. 84, pp. 85-91, 2012.

[6] B.W. Fehr, J.B. Gerrish, "Vision-guided row crop follower," *Applied Engineering in Agriculture,* vol. 11, no. 4, pp. 613-620, 1995.

[7] J.B. Gerrish, B.W. Fehr, G.R Van Ee, D.P. Welch, "Self-steering tractor guided by computer-vision," *Applied Engineering in Agriculture,* vol. 13, no. 5, pp. 559-563, 1997.

[8] J.B. Gerrish, T.C Surbrook, "Mobile robots in

agriculture," in *First International Conf. on Robotics and Intelligent Machines in Agriculture. ASAE*, St. Joseph,MI, 1984.

[9] J.B.Gerrish, G.C Stockman, "Image processing for path-finding in agricultural field operations," in *ASAE*, St. Joseph,MI, 1985.

[10] F. Wildrot, "Steering Attachment for Tractors". USA Patent 1506706, 1924.

[11] M.O'Connor, G. Elkaim, B. Parkinson, "Kinematic GPS for closed-loop control of farm and construction vehicles," in *ION-GPS*, Palm Springs, CA, 1996.

[12] J.F. Ried, S.W. Searcy, "Vision-based guidance of an agricultural tractor," *IEEE Control Systems,* vol. 7, no. 12, pp. 39-43, 1987.

[13] R.L. Schafer, R.E. Young, "An automatic guidance system for tractors," *Trans. ASAE,* vol. 22, no. 1, pp. 46-49, 1979.

[14] N. D Klassen, R.J. Wilson , N.J. Wilson, "Agricultural vehicle guidance sensor," in *ASAE*, St. Joseph, MI, 1993.

[15] M. Ollis, A. Stentz, "Vision-based perception for an automated harvester," in *IEEE International Conference on Intelligent Robots and Systems*, Piscataway, NJ, 1997.

[16] M. Ollis, A. Stentz, "First results in vision-based crop line tracking," in *IEEE*, Minneapolis, MN, 1996.

[17] Tech. Report: Economic Survey of Pakistan, 2011

[18] K.A. Quraishi, B. Akhtar,M. Aslam, Malick and A. Ali," *Socioeconomic Effect Of Industrialization On The Surrounding Rural Areas With Special Reference To Agriculture: A Case Study Of Islamabad District," Pakistan Journal Of Agricultural Sciences,* vol. 31, no. 3, pp. 236-240. (1994)